\documentclass[preprint,11pt,authoryear]{elsarticle}
\usepackage[utf8]{inputenc}
\usepackage[T1]{fontenc}

\usepackage{setspace}
\usepackage{lineno}

\usepackage{graphicx}
\usepackage{subcaption}
\usepackage{amssymb}
\usepackage{amsmath}
\usepackage{siunitx}
\usepackage{algorithmic}
\usepackage{algorithm2e} 
\usepackage{amsthm}
\usepackage{lscape}
\usepackage{verbatim}
\usepackage{color, soul}
\usepackage{xcolor}
\usepackage{tabularx, ragged2e, booktabs, caption, array, multirow, multicol}
\usepackage{csquotes}
\usepackage{mathtools}
\usepackage{dsfont}
\usepackage{listings}
\usepackage{latexsym}
\usepackage{epsfig}
\usepackage{float}
\usepackage{xspace}
\usepackage{tikz}
\usepackage{boldline}
\usepackage{lipsum}
\usepackage{bbm}
\usetikzlibrary{fit, calc}

\usepackage{hyperref}

\journal{Journal of Hydrology}

\makeatletter
\def\ps@pprintTitle{%
  \let\@oddhead\@empty
  \let\@evenhead\@empty
  \let\@oddfoot\@empty
  \let\@evenfoot\@oddfoot}
\makeatother
\begin{document}

\begin{frontmatter}

\title{QDeepGR4J: Quantile-based ensemble of deep learning and GR4J hybrid rainfall-runoff models for extreme flow prediction with uncertainty quantification}

%%%%%%%%%%%%%%%%%%%%%%%%%%%%%%%%%%%%%%%%%%%%%%%%%%%%%%%%%%%
% AUTHORS
%%%%%%%%%%%%%%%%%%%%%%%%%%%%%%%%%%%%%%%%%%%%%%%%%%%%%%%%%%%
\author[math,dare]{Arpit Kapoor\corref{cor1}}
\ead{arpit.kapoor@unsw.edu.au}

\author[math,dare]{Rohitash Chandra}
\ead{rohitash.chandra@unsw.edu.au} 

\cortext[cor1]{Corresponding author}

\address[math]{School of Mathematics and Statistics,
            University of New South Wales,
            Sydney, NSW 2052, Australia}

\address[dare]{Data Analytics for Resources and Environments,
            Australian Research Council—Industrial Transformation Training Centre,
            Sydney, NSW, Australia}

%%%%%%%%%%%%%%%%%%%%%%%%%%%%%%%%%%%%%%%%%%%%%%%%%%%%%%%%%%%
% ABSTRACT
%%%%%%%%%%%%%%%%%%%%%%%%%%%%%%%%%%%%%%%%%%%%%%%%%%%%%%%%%%%
\begin{abstract}
\begin{linenumbers}
    Conceptual rainfall-runoff models aid hydrologists and climate scientists in modelling streamflow to inform water management practices. Recent advances in deep learning have unravelled the potential for combining hydrological models with deep learning models for better interpretability and improved predictive performance.  In our previous work, we introduced DeepGR4J, which enhanced the GR4J conceptual rainfall-runoff model using a deep learning model to serve as a surrogate for the routing component. DeepGR4J had an improved rainfall-runoff prediction accuracy, particularly in arid catchments. Quantile regression models have been extensively used for quantifying uncertainty while aiding extreme value forecasting. In this paper, we extend DeepGR4J using a quantile regression-based ensemble learning framework to quantify uncertainty in streamflow prediction. We also leverage the uncertainty bounds to identify extreme flow events potentially leading to flooding. We further extend the model to multi-step streamflow predictions for uncertainty bounds. We design experiments for a detailed evaluation of the proposed framework using the CAMELS-Aus dataset. The results show that our proposed Quantile DeepGR4J framework improves the predictive accuracy and uncertainty interval quality (interval score) compared to baseline deep learning models. Furthermore, we carry out flood risk evaluation using Quantile DeepGR4J, and the results demonstrate its suitability as an early warning system.
\end{linenumbers}
\end{abstract}


\end{frontmatter}

\section*{Highlights}
\begin{itemize}
    
    \item Propose a quantile-based ensemble framework using hybrid rainfall-runoff models

    \item Feature uncertainty quantification for multistep streamflow prediction
    
    \item Results demonstrate that the framework improves predictive performance and uncertainty interval quality

     \item Present a qualitative measure of flood risk estimate from the predicted interval 

    \item Demonstrate the potential usefulness as an early-stage flood alert system.  
    
\end{itemize}

%%%%%%%%%%%%%%%%%%%%%%%%%%%%%%%%%%%%%%%%%%%%%%%%%%%%%%%%%%%
% INTRODUCTION
%%%%%%%%%%%%%%%%%%%%%%%%%%%%%%%%%%%%%%%%%%%%%%%%%%%%%%%%%%%
\section{Introduction}
\label{introduction}

Accurate prediction of extreme flows is vital due to the significant and diverse impacts these events impose on communities, ecosystems, and economies. Floods are among the most catastrophic natural disasters, resulting in substantial economic losses, environmental degradation, and tragic human fatalities  \citep{smith1994flood,halgamuge2017analysis,IPCC_AR6_WGII_2022,macmahon2015connecting}. Particularly severe floods can have widespread and lasting effects, with instances in New South Wales and Queensland alone causing minimum damages of about a million dollars in the last two decades, along with enduring economic and psychological consequences for those affected \cite{fernandez2015flooding}. The ability to reliably model and forecast such extreme events is therefore critical for effective water management and disaster response. Rainfall-runoff modelling is a key tool in modelling flooding events, and improving the accuracy of these models is essential to minimise the destructive consequences of floods that can help in better planning of evacuation \cite{lim2013review}. 

Conceptual models such as the  \textit{Australian Water Balance Model} (AWBM) \citep{boughton_australian_2004}, \textit{Génie Rural à 4 paramètres Journalier} model (GR4J) \citep{perrin_improvement_2003, perrin2007modeles}, and \textit{Sacramento} model \citep{burnash_nws_1995} have shown effectiveness in predicting streamflow, thereby aiding in the management of water resources and mitigating the impacts of climate change \citep{devia_review_2015, solomatine_216_2011, jehanzaib_comprehensive_2022, jaiswal_comparative_2020, hatmoko2020comparison}. However, a challenge with conceptual models is the data requirement for calibrating the model parameters, as well as poor performance on extreme event prediction. Physically-based hydrologic models rely on a mathematically idealised representation of underlying physical processes in the form of partial and ordinary differential equations \citep{abbott_introduction_1986, beven_changing_1989, beven_towards_2002, paniconi_physically_2015}. Due to their physical interpretation of the hydrologic processes, physically-based models do not require large volumes of meteorological or climate data to calibrate. While they overcome some limitations of the other modelling approaches, physically-based models suffer from scale-related problems due to complex underlying physics and decomposed architecture \cite{beven_changing_1989, jaiswal_comparative_2020}. 

Data-driven models have gained traction in streamflow prediction, including statistical time-series methods (e.g., ARIMA) \citep{valipour_long-term_2015, mishra_rainfall-runoff_2018}, neural networks \citep{tokar_rainfall-runoff_1999, dawson_artificial_1998} and deep learning models \cite{nearing2024global,chandra2024ensemble}. Data-driven models learn from empirical data rather than physical principles (such as physics-driven models) and have been demonstrated to outperform traditional approaches, especially in ungauged basins \citep{adnan_comparison_2021}. However, their lack of interpretability is a challenge \citep{samek_explainable_2019}.  Advances in explainable artificial intelligence (XAI) aim to address this issue \citep{montavon_methods_2018}, yet integrating physical processes into these models remains valuable for comprehensive understanding \citep{lees_hydrological_2022}.

Conceptual rainfall–runoff models are simple and interpretable, but structural error and equifinality often limit their transferability, especially under non-stationary or ungauged conditions \citep{beven_manifesto_2006,beven_future_1992}. In contrast, purely data-driven deep learning models can be data-hungry, lack physical consistency, and generalise poorly outside the training regime or during extremes, motivating physics-guided or hybrid approaches that embed process knowledge \citep{herath_hydrologically_2021, raissi_physics-informed_2019}. Hybrid modelling frameworks feature parsimony of conceptual or physics-based models with machine learning to enhance predictive accuracy while preserving interpretability \citep{bezenac_deep_2019, reichstein_deep_2019, razavi_coevolution_2022}. Various approaches have been proposed for hybridising environmental models, such as using machine learning for parameterisation of environmental/physical models \citep{beck_global-scale_2016}, modelling prediction errors in traditional approaches \citep{vandal_generating_2018}, replacing sub-processes in physics-based models with data-driven components \citep{bezenac_deep_2019}, and using machine learning as a surrogate to physical models \citep{camps-valls_physics-aware_2018, chevallier_neural_1998}. These methods often leverage gradient-free optimisation techniques, with evolutionary algorithms such as differential evolution, and particle swarm optimisation  \citep{guo2013novel, liu_automatic_2009, wang_genetic_1991}. Although these are effective for models with few parameters, gradient-based optimisation via backpropagation is more suitable for high-dimensional problems in deep learning \citep{ruder2016overview, krapu_gradient-based_2019}.

Quantile regression focuses on modelling conditional quantiles (such as the median) of a response variable, unlike linear regression, which estimates the conditional mean \citep{koenker1978regression}. This approach is particularly robust to outliers and useful for forecasting extremes \citep{portnoy1999extreme, cai2013extreme}. Recent applications have combined quantile regression with machine learning models. For example, \citet{taylor2000quantile} developed a quantile regression neural network, while \citet{wang2019probabilistic} integrated it with LSTM for load forecasting. \citet{pasche2022neural} used quantile regression and extreme value theory for flood risk forecasting. In hydrology, quantile regression has been used for downscaling precipitation \citep{bremnes2004probabilistic} and assessing forecast uncertainty in flood prediction \citep{weerts2011estimation, cai2013extreme}. 

In our previous work \citep{kapoor2023deepgr4j},   we presented a deep learning based hybrid model framework for rainfall-runoff modelling called DeepGR4J. The model utilised popular deep learning models (such as CNN and LSTM) to simulate the routing storage processes in the GR4J conceptual hydrologic model. A hierarchical optimisation framework was also presented that combines evolutionary optimisation for the hydrologic model with gradient-based optimisation for the deep learning models.
DeepGR4J showed considerable improvement in accuracy when compared to the baseline conceptual and machine learning models trained independently. However, the results also show that the DeepGR4J  suffers from poor performance in extreme flow regions. Therefore, DeepGR4J  needs to be adapted to make it suitable for predicting extremely high flows such as floods. Lastly, DeepGR4J does not address any uncertainty in model predictions, which is crucial for increasing the adoption and reliability of data-driven/hybrid modelling.

In this paper, we present an extension to the DeepGR4J model using quantile regression to provide additional predictions (quantiles) in an ensemble framework, which can be utilised to quantify the data uncertainty. We also present a qualitative measure of flood likelihood called \textit{flood risk indicator} utilising a  Generalised Extreme Value (GEV) distribution and the predicted uncertainty bounds from the quantile DeepGR4J framework. This approach aims to address the problem of extreme flows through a reliable early flood warning system. We test our framework on various stations from the CAMELS Australia (CAMELS-Aus) hydro-meteorological dataset and compare it with the baseline deep learning models. 
We also extend the model to a multi-step ahead prediction for long-term forecasting of streamflow with uncertainty intervals. 

The rest of the paper is organised as follows. Section 2 presents background on hydrological, deep learning methods and quantile regression. Section 3 presents our DeepGR4J-Extreme framework and model training scheme.  
Section 4 presents the experiment design and results, Section 5 discusses the results, and Section 6 concludes with future research directions.

%%%%%%%%%%%%%%%%%%%%%%%%%%%%%%%%%%%%%%%%%%%%%%%%%%%%%%%%%%%
% BACKGROUND
%%%%%%%%%%%%%%%%%%%%%%%%%%%%%%%%%%%%%%%%%%%%%%%%%%%%%%%%%%%

\section{Background}
\label{background}

\subsection{GR4J Hydrologic Model}
\label{gr4j}

\begin{figure}[htb]
    \centering
    \begin{subfigure}[b]{0.70\columnwidth}
      \includegraphics[width=0.95\linewidth]{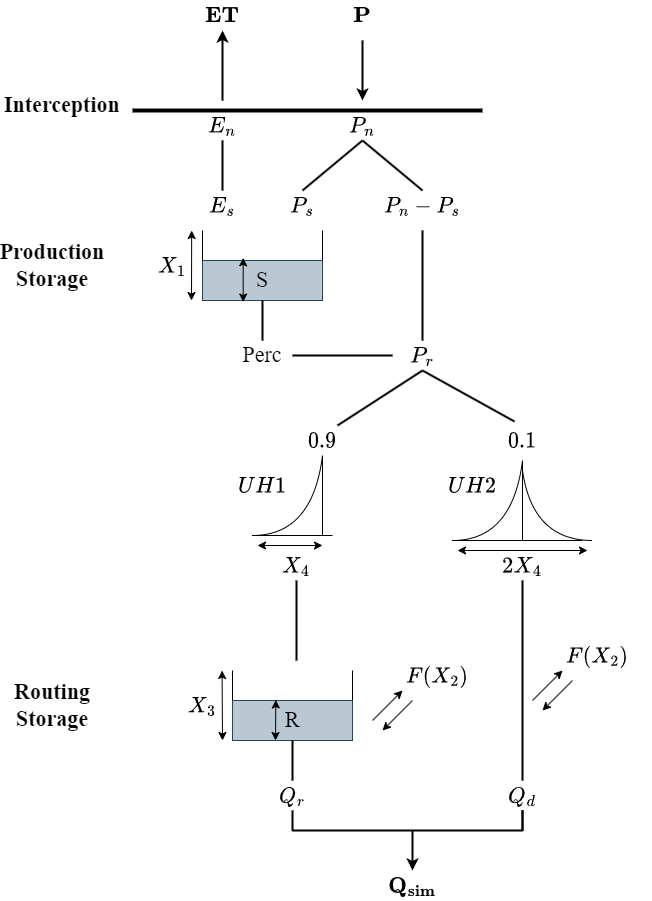}
    \end{subfigure}
    \caption{GR4J hydrologic model architecture}
    \label{fig:GR4J_arch}
\end{figure}

GR4J  \citep{perrin_improvement_2003, perrin2007modeles}  is a conceptual rainfall-runoff model designed to simulate daily streamflow in a catchment with water balance conditions (Figure \ref{fig:GR4J_arch}) . It relies on four tunable parameters: the maximal capacity of the production store ($X_1$), the catchment water exchange coefficient ($X_2$), the maximal routing reservoir capacity ($X_3$), and the unit hydrograph time base ($X_4$). At the core of GR4J are two storage components, including the \textit{production storage} representing stored soil moisture, and the \textit{routing storage} accounting for leakage and soil moisture exchange. The production storage regulates the allocation of precipitation between soil moisture recharge and direct runoff, whereas the routing storage represents the temporal delay and attenuation of runoff as it propagates through the catchment's hydrological pathways to the outlet.

At each time step $t$, we can compute the net precipitation ($P_n^{(t)}$) and evapotranspiration ($E_n^{(t)}$) in the production storage as:

\begin{eqnarray}
    P_{n}^{(t)} &=& \max(P^{(t)} - E^{(t)}, 0) \\
    E_{n}^{(t)} &=& \max(E^{(t)} - P^{(t)}, 0)
\end{eqnarray}

where $P^{(t)}$ and $E^{(t)}$ are the precipitation and evapotranspiration, at time step $t$. A portion of the net precipitation ($P_s^{(t)}$) enters the production store, while the remainder moves to the routing process. We can update the production store moisture component as follows:

\begin{equation}
    S^{(t)} = S^{(t-1)} + P_{s}^{(t)} - E_{s}^{(t)}
\end{equation}

We can compute the effective precipitation ($P_s^{(t)}$) and evapotranspiration ($E_s^{(t)}$) in the production store  based on the soil moisture level ($S^{(t-1)}$) and the storage capacity ($X_1$):

\begin{eqnarray}
    P_{s}^{(t)} &=& \frac{X_1 \left[1 - \left(\frac{S^{(t-1)}}{X_1}\right)^2\right] \tanh\left(\frac{P_{n}^{(t)}}{X_1}\right)}{1 + \frac{S^{(t-1)}}{X_1} \tanh\left(\frac{P_{n}^{(t)}}{X_1}\right)} \\
    E_{s}^{(t)} &=& \frac{S^{(t-1)}\left[2 - \frac{S^{(t-1)}}{X_1}\right] \tanh\left(\frac{E_{n}^{(t)}}{X_1}\right)}{1 + \left(1 - \frac{S^{(t-1)}}{X_1}\right) \tanh\left(\frac{E_{n}^{(t)}}{X_1}\right)}
\end{eqnarray}

We can then compute  the \textit{percolation} from the production store   as:

\begin{equation}
    Perc^{(t)} = S^{(t)} \left[1 - \left\{ 1 + \left(\frac{4}{9} \frac{S^{(t)}}{X_1}\right)^4\right\}^{-1/4}\right]
\end{equation}

This percolation flows to the \textit{routing storage}, which models how water moves through the catchment to generate streamflow. Inflow to the routing storage is split, with 90\% routed through a nonlinear reservoir and 10\% directed as quick flow via a secondary hydrograph. The routing storage also accounts for groundwater exchange, and the total runoff is a combination of the slow and quick flow components.

The GR4J model’s strength lies in its simple yet effective representation of soil moisture and runoff generation, particularly through the dynamics of its production store, which directly controls how precipitation is partitioned between storage, evapotranspiration, and runoff. This focus on water balance within the catchment, using only four parameters, makes GR4J well-suited for studies on catchment-scale hydrological processes.

\subsection{Quantile Regression for neural networks}

Quantile regression \cite{koenker2001quantile} is a statistical model for the conditional quantiles of the response variable (prediction) which provides a comprehensive understanding of the relationship between the input and prediction, as this enables us to quantify the uncertainty (aleatoric) present in data \citep{hao2007quantile}. Modelling different quantiles is particularly useful when the relationship between the input and the response varies across different quantiles.  Although quantile regression has traditionally been applied to models with fewer parameters, it has recently been integrated with deep neural networks to harness their flexibility and capacity to model complex, nonlinear relationships \citep{taylor2000quantile, cannon2011quantile, zhang2018improved}. Quantile regression can be implemented for neural networks for predicting the $\tau^{th}$ quantile of streamflow by minimising the tilted loss function, shown below:

\begin{equation}
    \mathcal{L}_\tau(\theta) = (\tau-1) \sum_{Q_i<\hat{Q}_i}\left(Q_i-\hat{Q}_i\right)+\tau \sum_{Q_i \geq \hat{Q}_i}\left(Q_i-\hat{Q}_i\right)
    \label{eq:quantreg}
\end{equation}

where $\theta$ refers to the neural network weights, $Q_i$ is the observed streamflow value for $i^{th}$ data sample, $\hat{Q}_i$ is the prediction. In this case, $\tau$ can take real values in the interval $[0, 1]$, and the quantile neural network model can be trained using the following optimisation problem: 

\begin{equation}
    \hat{\theta}_\tau = \underset{\theta \in \mathbb{R}}{\arg \min }\left\{ \mathcal{L}_\tau (\theta) \right\}
\end{equation}

where, $\hat{\theta}_\tau$ are the optimal neural network parameters for predicting the  $\tau^{th}$ quantile of the streamflow. In this study, we train deep learning models for quantiles $\tau = \{0.05, 0.50, 0.95\}$, which together define a 90\% confidence interval. In their recent work on quantile deep learning, \citet{cheung2024quantile} follow a similar choice of quantiles as this balances the ability to capture uncertainty with computational efficiency. This approach also aligns with common practice in hydrological forecasting, where a central estimate and bounds of a 90\% uncertainty interval are operationally useful. We can incorporate additional quantiles if a finer resolution of the uncertainty is required.

We use the standard backpropagation via the Adaptive moment estimation (Adam optimiser) \citep{kingma2014adam} for training the quantile-based neural network model. Adam features an adaptive learning rate computed individually for each neural network parameter using the first and second-order moments of the gradient, generally leading to faster yet sometimes poorer convergence compared to Stochastic Gradient Descent (SGD) with a fixed learning rate. This trade-off makes Adam particularly suitable in our setting, where rapid convergence, ease of use and stable training across heterogeneous parameters are essential.

\subsection{Evolutionary Algorithms for Hydrological Modelling}

Evolutionary algorithms have proven to be effective tools in calibrating conceptual hydrological models. In early studies,  \citet{franchini_use_1996} investigated various genetic algorithms (GA) for the calibration of rainfall-runoff models and highlighted their flexibility in optimising non-linear and multi-modal objectives. \citet{thyer1999probabilistic} demonstrated that probabilistic optimisation methods such as simulated annealing (SA) and shuffled complex evolution (SCE) are effective methods for navigating complex parameter spaces. \citep{wang_genetic_1991} demonstrated that evolutionary methods such as GA and SCE can also be successfully extended to distributed hydrological models. They also observed that the performance of GA approaches in calibrating distributed rainfall-runoff models may vary based on catchment attributes and objective functions. Furthermore, recent comparative studies highlight the adaptability and efficacy of various evolutionary approaches against modern data-driven and physics-based approaches for hydrological modelling \citep{tigkas2016comparative, kumar2019evaluation}. 

In addition to model calibration, evolutionary algorithms have also been used for data-driven modelling of hydrological processes, such as symbolic regression for rainfall-runoff modelling. Early studies showed that GP can discover interpretable model structures that reflect hydrological behaviour \citep{savic1999genetic, whigham2001modelling}. \cite{babovic2002rainfall} later built on this work by adding domain knowledge to the evolutionary process, which enhanced model adaptability for practical use cases. Multi-objective evolutionary algorithms (MOEAs) \cite{} are a class of evolutionary optimisation algorithms designed to handle problems with multiple conflicting objectives. \cite{tang2006effective} examined the efficacy and efficiency of various MOEA approaches for hydrologic model calibration based on computational efficiency, accuracy, and ease-of-use. In the past decade, researchers have also investigated the efficacy of combining evolutionary approaches with machine learning for hydrological applications. For example, \cite{sedki2009evolving} demonstrated the versatility of evolutionary computation in modern hydrological modelling by optimising neural network parameters for daily rainfall-runoff forecasts using a real-coded GA.

%%%%%%%%%%%%%%%%%%%%%%%%%%%%%%%%%%%%%%%%%%%%%%%%%%%%%%%%%%%
% Methodology
%%%%%%%%%%%%%%%%%%%%%%%%%%%%%%%%%%%%%%%%%%%%%%%%%%%%%%%%%%%

\begin{figure*}[!htb]
    \centering
    \includegraphics[width=1.0\linewidth]{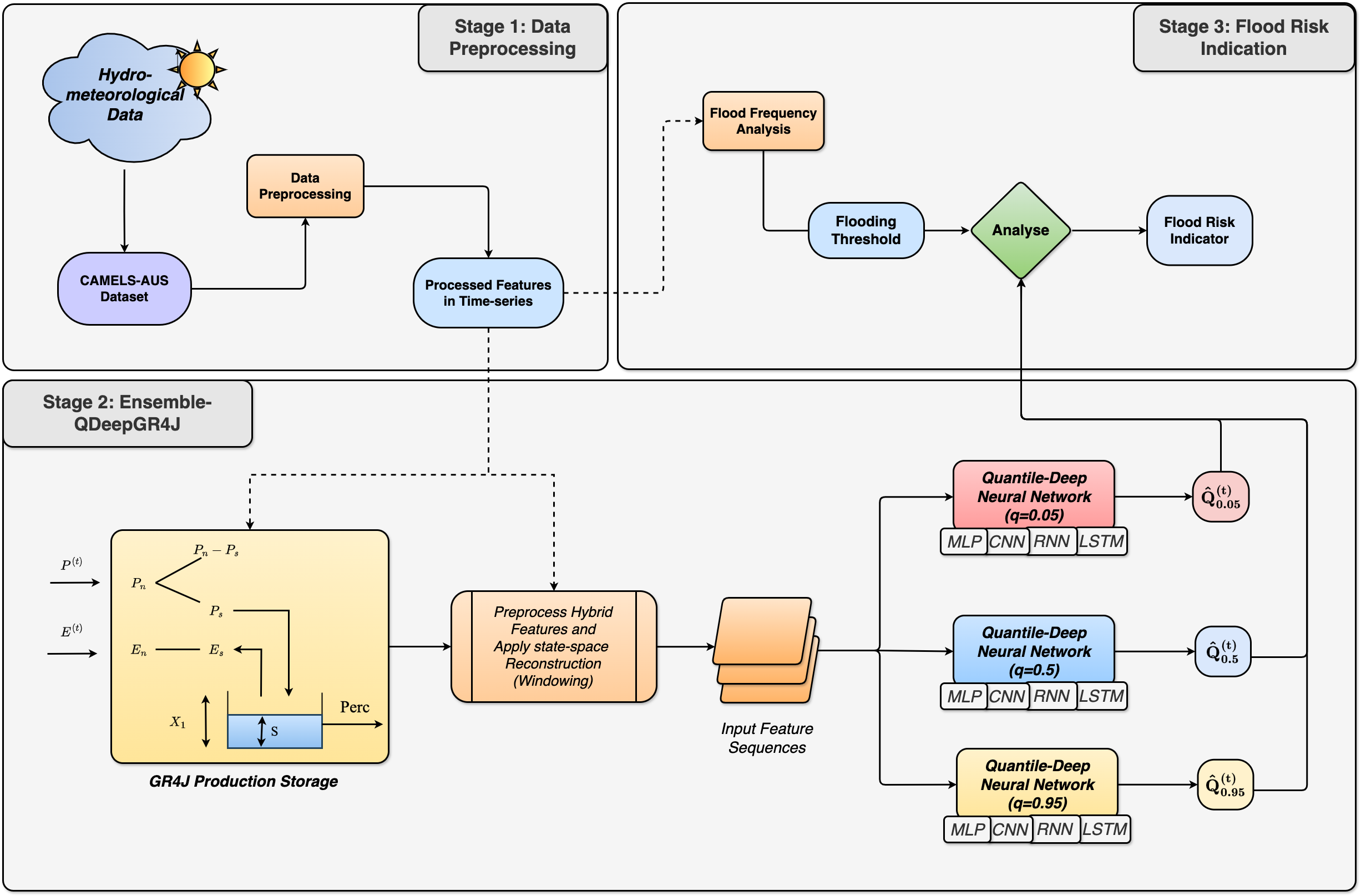}
    \caption{DeepGR4J-Extreme framework based on conditional ensembles catering to extreme values of streamflow}
    \label{fig:framework}
\end{figure*}

\begin{algorithm}[!htb]
    \SetAlgoLined
    \DontPrintSemicolon
    \small
    \KwData{Hydrological data for a gauged catchment}
    \KwResult{$X_1$ parameter and neural network parameters $\boldsymbol{\theta}$}
    \textbf{Stage 1:} Initialisation \\
    i. Define GR4J model $Q=gr4j(P, E; \delta)$ \\
    ii. IInitialise GR4J parameters $\delta = \{X_1, X_2, X_3, X_4\}$ \\
    iii. Define the ensemble of neural networks as $Q_\tau = g(\Tilde{\mathbf{X}}; \boldsymbol{\theta}_\tau)$ for $\tau \in \{0.05, 0.50, 0.95\}$\\
    v. IInitialise the neural network parameters, $\boldsymbol{\theta}_\tau$ \\
    \textbf{Stage 2:} Calibrate GR4J\\
    vi. Obtain optimal values of GR4J parameters as, $\hat{\delta}$ using differential evolution \\
    vii. Define production storage with optimised $X_1$ as $x = prod(P, E; \hat{X}_1)$\\
    \textbf{Stage 3:} Hybrid feature generation \\
    viii. Simulate features from the production storage: \\
    \For{$t=1,\ldots,T$}{
        1. Compute the feature from production storage: \\
        \hskip2.5em $\mathbf{x^{(t)}_{prod}} := prod(P^{(t)}, E^{(t)}; \hat{X}_1)$ \\
        2. Concatenate $\mathbf{x^{(t)}_{prod}}$ with meteorological features to obtain $\Tilde{\mathbf{X}}^{(t)}$
    }
    \textbf{Stage 4:} Quantile neural network training \\
    ix. Set neural network hyperparameters i.e. training epochs $N_{epoch}$, learning rate $\eta$\\
    x. Train the neural network models: \\
    \For{$\tau \in \{0.05, 0.50, 0.95\}$}{
        \For{$n=1,\ldots,N_{epoch}$}{
            \For{$t=1,\ldots,T$}{
                \For{$\tau\in\{0.05, 0.5, 0.95\}$}{
                    1. Obtain input and target pair: $(\Tilde{\mathbf{X}}, Q)$ \\
                    2. Predict streamflow quantile using the neural network, $\hat{Q}_\tau := g(\Tilde{\mathbf{X}}^{(t)}; \boldsymbol{\theta}_\tau)$\\
                    3. Compute loss $\mathcal{L}_\tau$ and gradients: $\Delta\boldsymbol{\theta} := \frac{\partial \mathcal{L}_\tau}{\partial\boldsymbol{\theta}}$ \\                
                    6. Update parameters: $\boldsymbol{\theta}_\tau := \boldsymbol{\theta}_\tau - \eta\Delta\boldsymbol{\theta}$\\
                 }
            }
        }
    }
    \caption{Hierarchical training of QDeepGR4J model}
    \label{alg:training}
\end{algorithm}

\section{Methodology}

\subsection{Data processing}
\label{sec:data_processing}

\begin{figure*}[htb]
    \centering
    \includegraphics[width=0.65\linewidth]{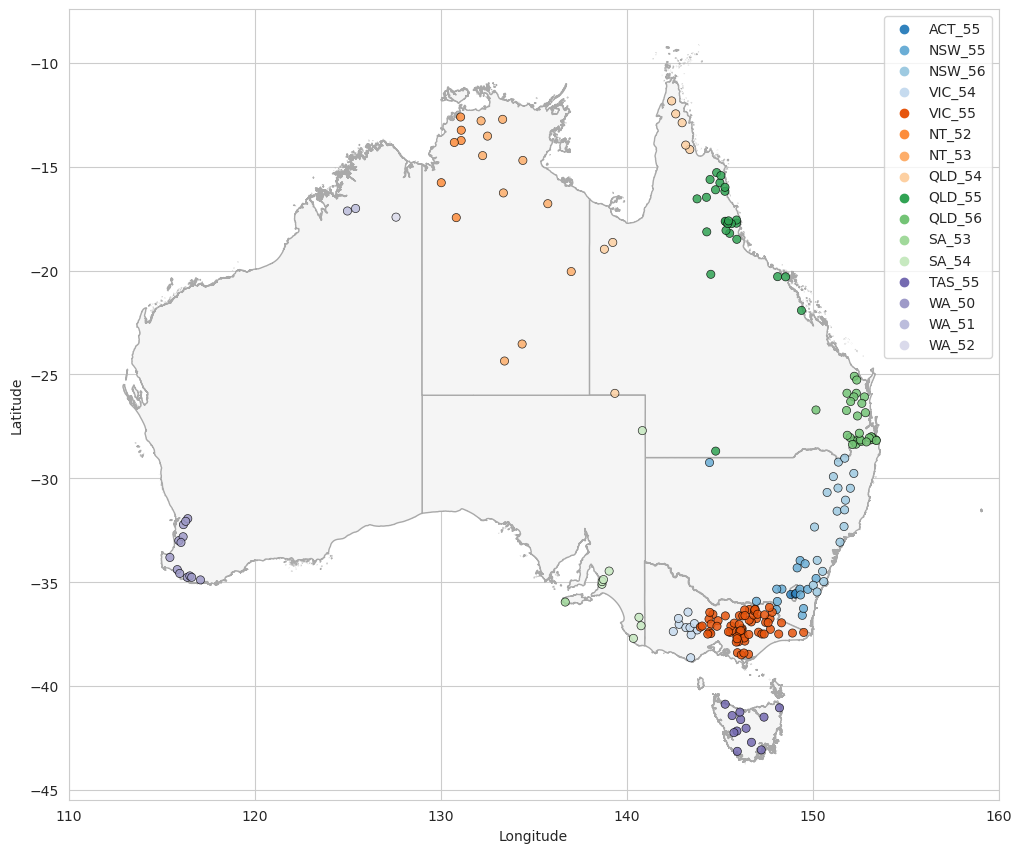}
    \caption{Stations lying in different regions across Australia}
    \label{fig:regions}
\end{figure*}

The Catchment Attributes and Meteorology for Large-sample Studies  (CAMELS) \cite{addor2017camels} dataset features hydrometeorological time series data for selected catchments across the United States. The CAMELS Australia  (CAMELS-AUS) is a region-specific instance that we utilise in our study, which includes hydrometeorological time-series data for 222 unregulated Australian catchments, covering streamflow, 12 climate variables, and 134 catchment attributes related to geology, soil, topography, and others \citep{fowler_camels-aus_2021}. The dataset spans over four decades for most catchments, offering valuable insights into arid-zone hydrology. Figure \ref{fig:regions} shows the location of stations in the dataset, colour-coded based on the state and map zone. It should be noted that most of the stations are located in the southeast region of Australia, and very limited data is available for stations on the west coast of Australia. We processed the data using the `camels-aus-py` python package \footnote{\url{https://github.com/csiro-hydroinformatics/camels-aus-py}}, developed by CSIRO \footnote{Commonwealth Scientific and Industrial Research Organisation, Australian Government: https://www.csiro.au/en/}. We focus on the period from 1980 to 2014, splitting 60\% for model training and 40\% for testing as used in previous works \citep{kapoor2023deepgr4j}.

We reconstruct the multivariate time series data using a windowing approach inspired by Taken’s Theorem, with a window size of $\alpha = 7$ for the respective models. We highlight that some stations in the dataset contain substantial gaps in the time-series variables of interest %(see Table \ref{tab:features}). 
We discard any station where more than 10\% of the time steps are missing for at least one variable. For the remaining stations, missing values are imputed using linear interpolation.  We standardise both input and output data using z-score normalisation \citep{abdi2010normalizing} before model training. The complete data processing pipeline, including an updated version of the \textit{camels-aus-py} package, is available in our public repository \footnote{GitHub link: \url{https://github.com/DARE-ML/DeepGR4J-Extremes}}

\subsection{Ensemble QDeepGR4J hybrid rainfall-runoff model}

In our previous work, we introduced DeepGR4J \citep{kapoor2023deepgr4j}, a hybrid rainfall-runoff model that builds upon the traditional GR4J (a daily lumped rainfall-runoff model) \citep{perrin2007modeles}. DeepG4RJ incorporates a deep learning model in GR4J  to enhance hydrological predictions. DeepG4RJ combines the production storage components of the GR4J model and deep learning models such as  Convolutional Neural Networks (CNN) \citep{lecun1995convolutional,lecun_gradient-based_1998} and Long Short-Term Memory (LSTM) \citep{hochreiter_long_1997,hochreiter_vanishing_1998} recurrent neural networks (RNN).

We extend the DeepGR4J rainfall-runoff model using quantile regression to predict specific streamflow quantiles. We refer to this model as \textit{Quantile-DeepGR4J} (QDeepGR4J), which employs a multi-stage model training approach combining differential evolution for calibrating the GR4J parameters and gradient descent for training the neural network model. We incorporate the tilted loss function (Equation \ref{eq:quantreg}) for the training of quantile regression-based deep learning models that include CNN, LSTM, and RNN. We also include  MLP for comparison of results. 

Figure \ref{fig:framework} provides an overview of the  QDeepGR4J framework for computing the \textit{flood risk indicator} values using the ensemble quantile-based DeepGR4J hybrid rainfall-runoff model. We preprocess the hydro-meteorological time-series dataset in the first stage of the framework. QDeepGR4J is a catchment-specific hybrid rainfall-runoff model. Stage 2 consists of training the ensemble quantile-based DeepGR4J model on the processed data. During this stage, we also compute the multi-step predictions for streamflow with uncertainty bounds using the trained model.  As shown in Stage 2 (Figure \ref{fig:framework}),  the model begins by utilising hydro-meteorological data, which includes precipitation $(P^{(t)})$ and evapotranspiration $(E^{(t)})$ time series, serving as input to the GR4J production storage component. This component processes the data to simulate storage dynamics, which includes components like production $( P_n )$, inter-storage transfer $( P_s )$, and percolation (\textit{Perc}), effectively estimating intermediate variables that capture hydrological states in the catchment. These outputs are passed through a hybrid pre-processing step involving state-space reconstruction and windowing, which reshapes the data into \textit{reconstructed features} suitable for deep learning models for time-series data. %got this paragraph from next subsec

In the subsequent stage, we input the reconstructed feature data into a machine learning model, such as MLP, CNN, RNN, and LSTM. In our previous work, \citet{kapoor2023deepgr4j} demonstrated that the CNN model outperformed LSTM in most cases for the standard DeepGR4J approach. However, given that the learning problem has shifted from a mean regression to a quantile regression problem, we evaluate additional architectures in this study. We train an ensemble of three QDeepGR4J models: $\text{QDeepGR4J}_{\tau=0.05}$, $\text{QDeepGR4J}_{\tau=0.50}$, and $\text{QDeepGR4J}_{\tau=0.95}$, corresponding to lower, median and upper quantiles of streamflow, respectively. The predicted quantiles of the streamflow at time-step $t$ are represented by $\hat{Q}_{0.05}^{(t)}, \hat{Q}_{0.5}^{(t)}$ and $\hat{Q}_{0.95}^{(t)}$. Therefore, using the quantile predictions, we can construct a 90\% confidence interval for streamflow predictions. Finally, in Stage 3, we use the predicted uncertainty bounds for the streamflow to generate the \textit{flood risk indicator} values. 

Details of the framework are discussed in the following sections. As shown in Algorithm \ref{alg:training}, the training algorithm for QDeepGR4J starts with initialisation (Stage 1), where the GR4J model, defined by parameters $(\delta = \{X_1, X_2, X_3, X_4\})$ and two inputs: precipitation ($P^{(t)}$) and evapotranspiration ($E^{(t)}$). We define the neural network model (eg, CNN, LSTM, RNN, and MLP) $g(\Tilde{\mathbf{X}};\boldsymbol{\theta})$ by input features $(\Tilde{\mathbf{X}})$ and model parameters including weights and biases $( \boldsymbol{\theta})$. In Stage 2, we calibrate  GR4J parameters using Differential Evolution to obtain optimal values of $(\hat{\delta})$. The calibrated GR4J model supports efficient simulation of catchment storage dynamics through the production storage $(prod(P, E; \hat{X}_1))$. Once calibrated, in Stage 3 we generate the hybrid features (as shown in Figure \ref{fig:framework})  by combining the production storage outputs $(P_n, E_n, P_s, Perc)$ with meteorological inputs $(P, E, T_{min}, T_{max}, vprp)$, forming an enhanced feature set $(\Tilde{\mathbf{x}}^{(t)})$ that incorporates hydrological and meteorological dynamics. We convert the time series data from the hybrid feature set to input sequences $(\Tilde{\mathbf{X}}^{(t)})$ using a window size of $\alpha$;

\begin{equation}
    \Tilde{\mathbf{x}}^{(t)} = \begin{bmatrix}
        P^{(t)},  
        E^{(t)}, 
        T_{min}^{(t)}, T_{max}^{(t)}, 
        vprp^{(t)},
        P_{n}^{(t)}, E_{n}^{(t)}, P_{s}^{(t)}, {Perc}^{(t)}
    \end{bmatrix}
\end{equation}

\begin{equation}
    \Tilde{\mathbf{X}}^{(t)} = \begin{bmatrix}
        {\Tilde{\mathbf{x}}^{(t+1)}} \vspace{2mm} \\
        {\Tilde{\mathbf{x}}^{(t+2)}} \vspace{2mm} \\
        \vdots \vspace{2mm} \\
        {\Tilde{\mathbf{x}}^{(t+\alpha)}}
    \end{bmatrix}
\end{equation}

We then use the hybrid time series dataset to train the respective deep learning models via quantile regression, which allows for targeted predictions of streamflow quantiles $Q_\tau$ at various levels (e.g., $\tau = 0.05, 0.5, 0.95$ for low, median, and high flow conditions). The ensemble quantile-based deep learning models individually predict the lower bound, upper bound and the median value of the 90\% confidence interval for the streamflow. The model training proceeds by calculating a quantile-specific loss $( \mathcal{L}_\tau )$ for each output unit, computing gradients $( \Delta\boldsymbol{\theta})$ for the neural network model, and updating parameters $(\boldsymbol{\theta})$ via gradient descent. This quantile-focused training process enables DeepGR4J to provide robust predictions across the flow spectrum, making it well-suited for managing variable and extreme hydrological events with a clear accounting of uncertainties. It refines the model's ability to handle diverse hydrological conditions, particularly effectively capturing extremes and associated uncertainties. Additionally, we train the QDeepGR4J ensemble models for multi-step-ahead streamflow prediction, specifying a 3-day forecast horizon. \citet{patel_enhancing_2024} demonstrated that 3-day lead times yield excellent results for flood prediction. This motivated a 3-day forecast horizon for streamflow quantiles, along with the operational relevance for dam pre-release and community warning, especially in the Australian context. Lastly, we used Adam-based model training configured with a learning rate of 0.001 and moment parameters ($\beta_1 = 0.89$ and $\beta_2 = 0.97$).

\subsection{Flood Risk Indicator}

\label{sec:flood_risk_indicator}

We compute the flood risk indicator as a qualitative label of flooding risk, based on the streamflow predictions. As demonstrated in previous work \citep{chandra2024ensemble}, the flood risk indicator can be computed as a function of predicted streamflow and the flood threshold $\gamma$. Although we adopt the same definition of the flood risk indicator, we update the approach to a Generalised Extreme Value (GEV) \cite{haan2006extreme} distribution for computing $\gamma$.

The GEV distribution is ideal for modelling the distribution of block maxima. In hydrology, it is often used to model the extreme precipitation and streamflow events \cite{}. It consists of three extreme value distributions, including  Gumbel, Fréchet, and Weibull, that are connected via a shape parameter which governs the tail behaviour.  We can estimate the upper quantiles of flow associated with rare flooding events by fitting the GEV distribution to observed annual maximum streamflow data. This provides a well-defined approach to estimate a flood threshold for the particular catchment. We define the GEV distribution  by the following probability density function:

\begin{equation}
    f(u; \zeta) = \begin{cases}
        \exp(-(1-\zeta u)^{1/\zeta})(1-\zeta u)^{1/c-1} & \text{if } \zeta \neq 0 \\
        \exp(-\exp(-u))\exp(-u) & \text{if } \zeta = 0
    \end{cases}
    \label{eq:gev}
\end{equation}

where $u = (x - \mu)/\sigma$ with $\mu$ and $\sigma$ as the location and scale parameters, and $\zeta$ is the shape parameter that follows $-\infty < u \leq 1/\zeta$ if $\zeta>0$ and $1/\zeta \leq u<\infty$ if $\zeta<0$. We fit the GEV parameters (ie, $\zeta, \mu, \sigma$) using the observed annual maximum flow data. The flood threshold $\gamma$ is then computed as $\gamma=F^{-1}(p; \zeta, \mu, \sigma)$ where $F^{-1}(\cdot)$ is the inverse of the cumulative distribution function (CDF) of the GEV. For a once in $k$ year flood, we set the value of $p=(1-1/k)$. So, for once in a 5-year flood, $p=0.80$. We compute the flood risk indicator  as:

\begin{eqnarray}
     FRI = \begin{cases} 
              \text{High} &  \max(\hat{Q}_{0.05}) > \gamma \\
              \text{Moderate} & \max(\hat{Q}_{0.50}) > \gamma \\
              \text{Low} & \max(\hat{Q}_{0.95}) > \gamma \\
              \text{Unlikely} & Otherwise
           \end{cases}
    \label{eq:flood_risk_indicator}
\end{eqnarray}

where, $\hat{Q}_{\tau}$ is the $\tau$ percentile of the streamflow predicted by the QDeepGR4J$_{\tau}$ model over the forecast horizon. The $\max(\cdot)$ function is used to compute the maximum value of predicted flow within the forecast horizon.

\subsection{Evaluation strategy}

{
 We evaluate the median value ($\tau=0.50$) predictions using the Root Mean Squared Error (RMSE) and Nash–Sutcliffe Model Efficiency Coefficient (NSE) scores:
\begin{eqnarray} 
    RMSE &=& \sqrt{\frac{\sum_{m=1}^{M}\sum_{t=1}^{T}(\hat{Q}^{(mt)}-Q^{(mt)})^2} {M\times T}} \label{eq:rmse}\\
    NSE &=& 1 - \frac{\sum_{m=1}^{M}\sum_{t=1}^{T}(\hat{Q}^{(mt)}-Q^{(mt)})^2}{\sum_{m=1}^{M}\sum_{t=1}^{T}(Q^{(mt)}-\bar{Q^{(mt)}})^2} \label{eq:nse}
\end{eqnarray}
where $M$ is the number of sequences in the data and $T$ is the length of the prediction horizon. Additionally, we calculate the interval score (IS) to quantitatively evaluate the quality of the predicted confidence interval. We compute the  interval score as follows: 
\begin{eqnarray}
    IS &=& (U - L) + \frac{2}{\delta}(L - Q) * \mathbbm{1}(Q < L) \nonumber \\
    &&+ \frac{2}{\delta}(Q - U) * \mathbbm
    {1}(Q > U)
    \label{eq:interval_score}
\end{eqnarray}
where, $\mathbbm{1}$ denotes the indicator function, $U$ and $L$ are the predicted upper $(\hat{Q}_{0.95})$ and lower $(\hat{Q}_{0.05})$ bounds, respectively. $Q$ is the observed value of streamflow, and $\delta=0.1$ corresponding to the 90\% confidence interval.  A lower interval score is desirable since it penalises the interval width as well as the number of observations lying outside the predicted interval.}
%%%%%%%%%%%%%%%%%%%%%%%%%%%%%%%%%%%%%%%%%%%%%%%%%%%%%%%%%%%
% Experiments and Results
%%%%%%%%%%%%%%%%%%%%%%%%%%%%%%%%%%%%%%%%%%%%%%%%%%%%%%%%%%%
\section{Experiments and Results}

\subsection{Experiment Design}

We design experiments to compare the different models for streamflow prediction, organised as follows:

\begin{enumerate}
    
    \item We compare and identify the most suitable neural network model for the QDeepGR4J model based on the predictive performance.

    \item We evaluate the performance of the best machine learning model in the QDeepGR4J model configuration by comparing LSTM and CNN models for selected catchments across different states.

    \item We compute and evaluate the \textit{flood risk indicator} using the best-performing QDeepGR4J ensemble. 

\end{enumerate}

\subsection{Evaluation of neural network models}

\begin{figure*}[htb]
    \centering
    \begin{subfigure}[b]{0.49\textwidth}
        \includegraphics[width=\linewidth]{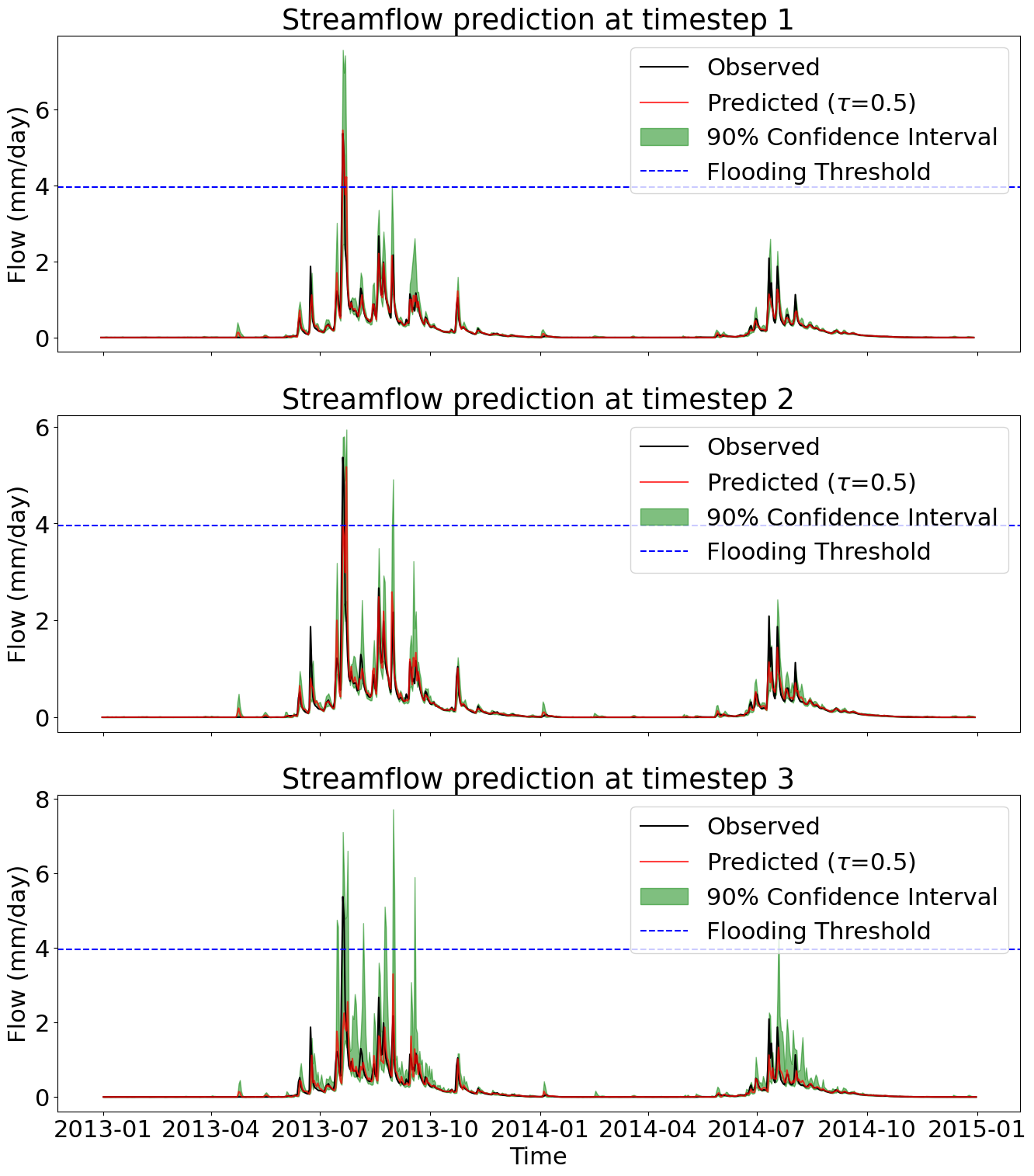}
        \caption{A5130501}
    \end{subfigure}
    \begin{subfigure}[b]{0.49\textwidth}
        \includegraphics[width=\linewidth]{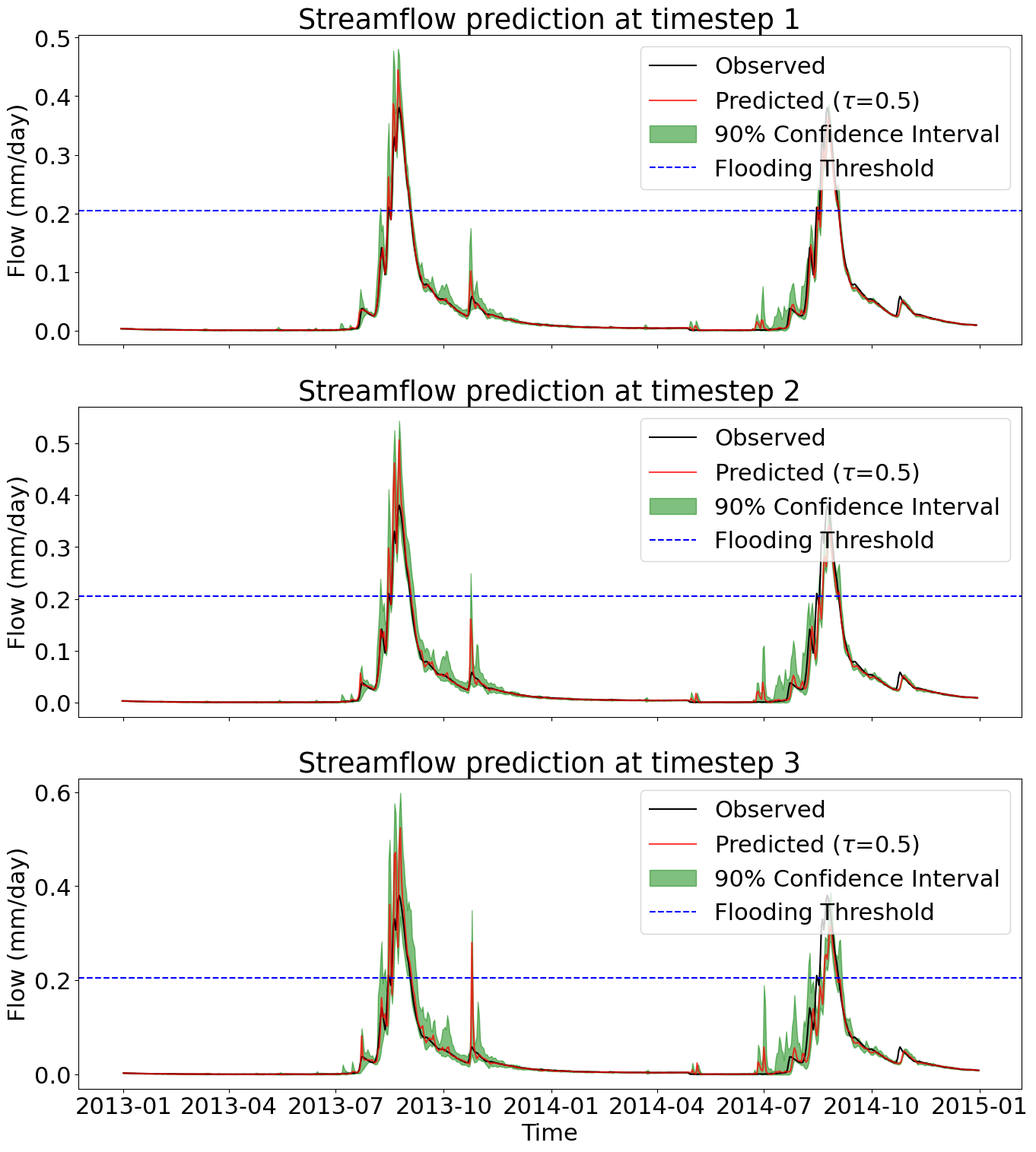}
        \caption{A2390523}
    \end{subfigure}
    \caption{QDeepGR4J-LSTM predictions for two stations located in South Australia}
    \label{fig:qdeepgr4j_preds}
\end{figure*}

We compare the performance of four neural network architectures using the QDeepGR4J ensemble framework, which includes CNN, vanilla RNN, LSTM, and MLP. We train the quantile-based ensembles (ie, $\tau \in {0.05, 0.50, 0.95}$) for each architecture separately. Table \ref{tab:arch_res} presents the performance of different QDeepGR4J model configurations for all stations in the South Australia (SA) region. The results show that in the case of median value prediction LSTM-based architecture has the best performance, followed by the MLP and CNN-based architectures. The simple RNN-based architecture shows the worst median value prediction performance. These results are consistent between the training and test datasets. The interval score results also show that the LSTM-based model yields a superior performance in capturing the desired uncertainty in the predictions when compared to the other architectures. However, we observe that the uncertainty quantification offered by MLP is the worst, and it is outperformed by the RNN, LSTM and CNN models. Overall, we observe that the LSTM-based model is the most suited architecture for the quantile-based DeepGR4J model, with significantly better performance in terms of the relevance of the predicted quantiles. 

\begin{table}[!htb]
    \centering
    \small
    \begin{tabular}{lrrrrrr}
        \toprule
         & \multicolumn{2}{r}{RMSE $(\tau=0.5)$} & \multicolumn{2}{r}{NSE $(\tau=0.5)$} & \multicolumn{2}{r}{Interval Score} \\
        Model & Train & Test & Train & Test & Train & Test\\
        \midrule
        MLP & 0.5616 & 0.4131 & 0.6499 & 0.6160 & 1.4220 & 0.9338 \\
        RNN & 0.6684 & 0.4716 & 0.3386 & 0.3551 & 0.9272 & 0.6244 \\
        CNN & 0.5451 & 0.3979 & 0.6129 & 0.5892 & 0.4830 & 0.4247 \\
        LSTM & 0.4792 & 0.3829 & 0.7775 & 0.6650 & 0.3679 & 0.4198 \\
        \bottomrule
    \end{tabular}
    \caption{Streamflow prediction performance of various QDeepGR4J model architectures on all stations in South Australia (SA)}
    \label{tab:arch_res}
\end{table}

\begin{table*}[!htb]
    \centering
    \small
    \begin{tabular}{llrrrrrr}
    \toprule
        &  & \multicolumn{2}{r}{RMSE $(\tau=0.5)$} & \multicolumn{2}{r}{NSE $(\tau=0.5)$} & \multicolumn{2}{r}{Interval Score} \\
        State & Model & Train & Test & Train & Test & Train & Test \\
        \midrule
        
        NSW & CNN           & 3.9731 & 2.8053 & 0.2267 & 0.3759 & 4.9681 & 3.9931 \\
           & LSTM           & 4.0016 & 2.8321 & 0.2172 & 0.3674 & 5.2591 & 4.0582 \\
           & DeepGR4J-CNN  & 3.8972 & 2.7564 & 0.2568 & 0.3959 & 4.4299 & 3.8994 \\
           & DeepGR4J-LSTM &\textbf{ 3.3881} & \textbf{2.4840} & \textbf{0.4462} & \textbf{0.4557} & \textbf{3.2633} & \textbf{3.6984} \\
        \midrule   
        
        NT & CNN            & 2.4051 & 3.1235 & 0.5224 & 0.5408 & 2.7625 & 4.0243 \\
           & LSTM           & 2.4255 & 3.1549 & 0.5084 & 0.5358 & 2.8767 & 4.3802 \\
           & DeepGR4J-CNN   & 2.2554 & 2.9391 & 0.5857 & 0.5994 & 2.5715 & 3.8614 \\
           & DeepGR4J-LSTM  & \textbf{1.8756} & \textbf{2.7486} & \textbf{0.7357} & \textbf{0.6568} & \textbf{1.7389} & \textbf{3.8564} \\
        \midrule
        
        QLD & CNN           & 8.0515 & 6.8162 & 0.4868 & 0.5287 & 11.8666 & 11.9243 \\
           & LSTM           & 8.1678 & 6.8828 & 0.4639 & 0.5165 & 12.1476 & 11.9032 \\
           & DeepGR4J-CNN   & 7.7990 & 6.6381 & 0.5170 & 0.5523 & 10.5873 & \textbf{11.4625} \\
           & DeepGR4J-LSTM  & \textbf{6.6927} & \textbf{6.3671} & \textbf{0.6373} & \textbf{0.5907} &  \textbf{8.0509} & 12.1210 \\
       \midrule
        
        SA & CNN            & 0.7072 & 0.5220 & 0.4880 & 0.4579 & 0.9177 & 0.7043 \\
           & LSTM           & 0.7303 & 0.5301 & 0.4824 & 0.4660 & 0.9257 & 0.6982 \\
           & DeepGR4J-CNN   & 0.6652 & 0.4976 & 0.5365 & 0.5076 & 0.8133 & 0.6639 \\
           & DeepGR4J-LSTM  & \textbf{0.6017} & \textbf{0.4829} & \textbf{0.6878} & \textbf{0.5649} & \textbf{0.5829} & \textbf{0.6394} \\
        \midrule
        
        TAS & CNN           & 2.3087 & 2.2435 & 0.6324 & 0.6503 & 5.3506 & 5.3200 \\
           & LSTM           & 2.2812 & 2.2221 & 0.6386 & 0.6528 & 5.0784 & 5.1561 \\
           & DeepGR4J-CNN   & 2.1977 & 2.1605 & 0.6720 & 0.6822 & 4.8403 & 5.1138 \\
           & DeepGR4J-LSTM  & \textbf{2.0910} & \textbf{2.0702} & \textbf{0.7239} & \textbf{0.7159} & \textbf{4.3913} & \textbf{4.9004} \\
        \midrule
        
        VIC & CNN           & 1.8210 & 1.2320 & 0.6778 & 0.6615 & 2.3859 & 2.0828 \\
           & LSTM           & 1.8117 & 1.2191 & 0.6775 & 0.6699 & 2.2291 & 1.9660 \\
           & DeepGR4J-CNN   & 1.7583 & 1.1919 & 0.7057 & 0.6885 & 2.1734 & 1.9674 \\
           & DeepGR4J-LSTM  & \textbf{1.6363} & \textbf{1.1106} & \textbf{0.7482} & \textbf{0.7313} & \textbf{1.7108} & \textbf{1.7884} \\
        \midrule
        
        WA & CNN            & 1.2250 & 1.8153 & 0.6060 & 0.6098 & 1.4093 & 1.5785 \\
           & LSTM           & 1.1750 & 1.7810 & 0.6294 & 0.6265 & 1.3238 & 1.5947 \\
           & DeepGR4J-CNN   & 1.1324 & 1.7477 & 0.6563 & 0.6462 & 1.2079 & \textbf{1.4832} \\
           & DeepGR4J-LSTM  & \textbf{0.9922} & \textbf{1.6344} & \textbf{0.7310} & \textbf{0.6900} & \textbf{0.8538} & 1.5692 \\
    
    \bottomrule
    \end{tabular}
    \caption{Comparison of Ensemble Quantile-based DeepGR4J (LSTM \& CNN) with baseline Ensemble Quantile-based Deep Learning models (LSTM \& CNN)}
    \label{tab:top5_res}
\end{table*}

Figure \ref{fig:qdeepgr4j_preds} presents the time series of observed streamflow along with the predicted quantiles for two randomly selected stations in the SA region. We present the results for the LSTM-based quantile-based ensemble DeepGR4J model over the three time steps in the prediction horizon. The green region corresponds to the 90\% confidence interval based on the predicted $5^{th}$ and $95^{th}$ percentiles. We can observe that the LSTM-based QDeepGR4J ensemble effectively captures the uncertainty in the streamflow prediction for all three timesteps, with a slight increase in the uncertainty bounds for time steps 2 and 3. We also notice that for some peaks, the model overestimates the upper bound significantly, especially for the Station ID - A5130501. 

\subsection{Evaluation across multiple regions}

\begin{figure}[htb]
    \centering
    \begin{subfigure}[b]{0.49\textwidth}
        \includegraphics[width=\linewidth]{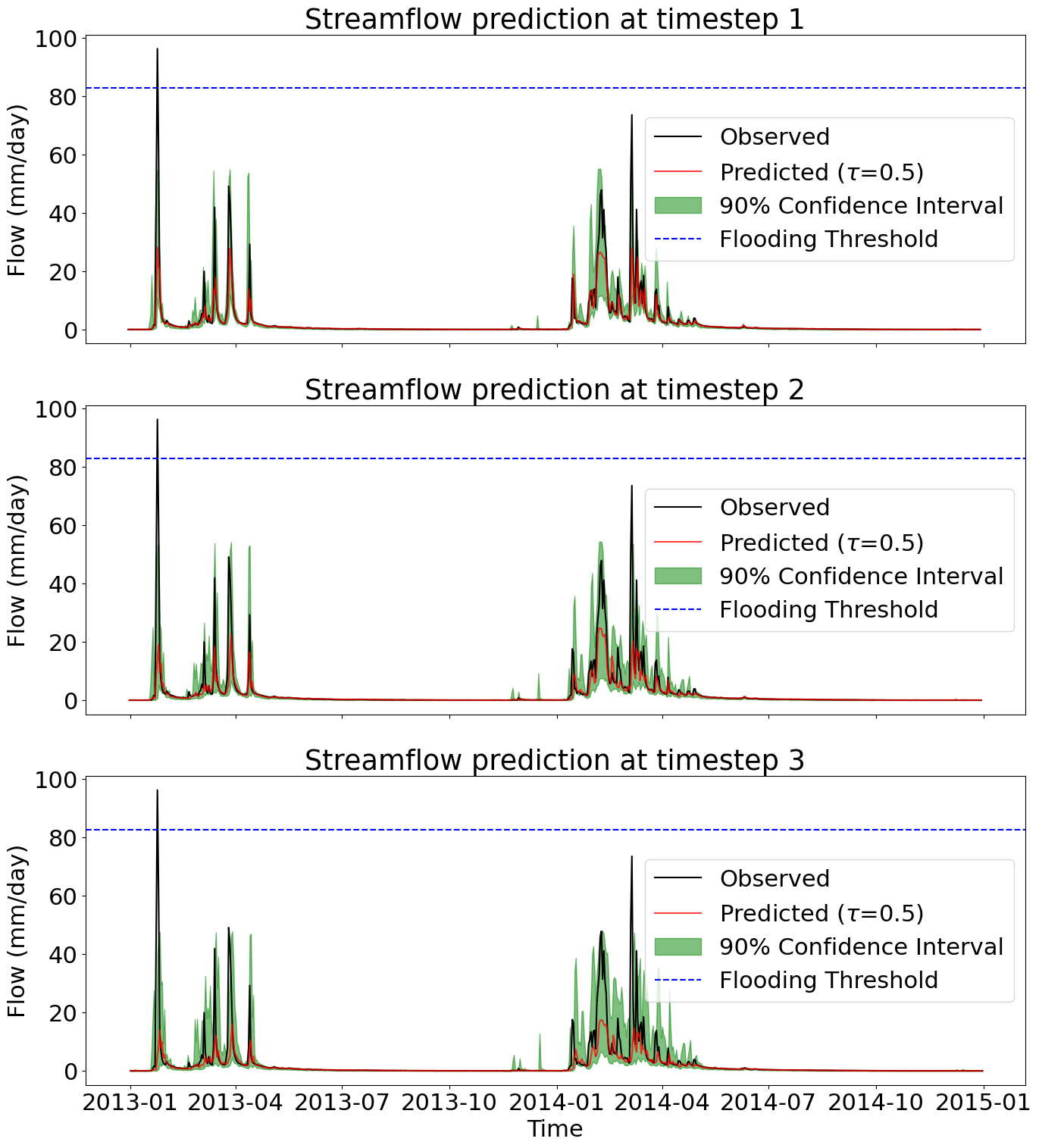}
        \caption{LSTM}
    \end{subfigure}
    \begin{subfigure}[b]{0.49\textwidth}
        \includegraphics[width=\linewidth]{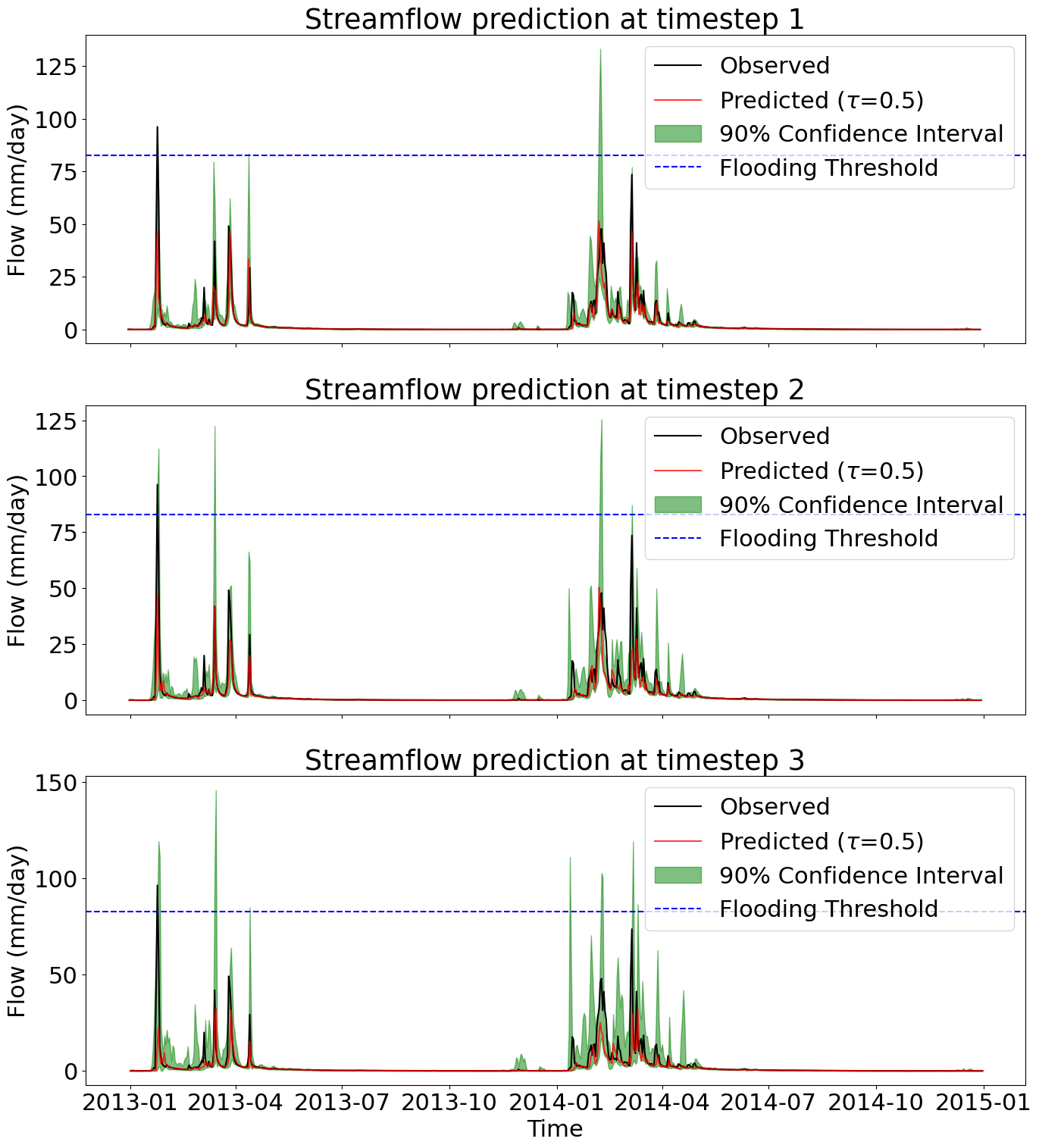}
        \caption{QDeepGR4J-LSTM}
    \end{subfigure}
    \caption{Comparison of streamflow quantile predictions from quantile-based LSTM ensemble and QDeepGR4J-LSTM ensemble for Pascoe River at Fall Creek station (102101A).}
    \label{fig:preds_comparison}
\end{figure}

In the previous experiment, we identified the LSTM as the most effective architecture for the ensemble QDeepGR4J model in terms of median predictive performance (RMSE and NSE), as well as the interval score. \citet{chandra2024ensemble} demonstrated that their LSTM-based quantile ensemble model is the most effective in capturing extreme flow behaviour compared to the other architectures. However, in our previous work \citep{kapoor2023deepgr4j}, we observed that the CNN-based hybrid model outperformed the LSTM-based model in single-step-ahead prediction (mean-value). Therefore, we evaluate the performance of our hybrid ensemble models (DeepGR4J-CNN and DeepGR4J-LSTM) against the baseline deep learning counterparts (CNN and LSTM). To ensure a fair comparison, we use the same deep learning model architectures for both hybrid and baseline models. Since catchments with high runoff ratios are more likely to experience extreme flow events, we chose five stations from each state that have the greatest runoff ratios. Since the Australian Capital Territory (ACT) covers a small region with only three available stations, we do not evaluate models for ACT stations for this experiment.  

Table \ref{tab:top5_res} presents the average performance metrics across the selected stations in each state. The results show that in all seven states, the LSTM-based QDeepGR4J ensemble demonstrates the best performance in median value prediction, followed by the CNN-based QDeepGR4J ensemble; i.e., in terms of RMSE and NSE scores of train and test sets. We observe that hybridisation provides a considerable improvement in NSE and RMSE performance for train and test sets in both LSTM and CNN-based ensembles. In the case of interval score performance, we observe that the LSTM-based QDeepGR4J model yields the lowest values of the train set in all states. However, in the case of the test set, we observe that the CNN-based QDeepGR4J model outperforms the LSTM-based counterpart for two states (QLD, WA). Note that RMSE is scale-dependent, whereas NSE is normalised by the variance of the observed flows in the evaluation period. As a result, when the test split exhibits larger flow variability (e.g., inclusion of high-flow events), the model can achieve a higher NSE on the test set while also exhibiting a larger RMSE in absolute units, as observed in the case of the Northern Territory (NT) state for the DeepGR4J-CNN ensemble. In such cases, we assign higher precedence to the NSE score as it presents a normalised score representing the amount of variance captured by the model. Figure \ref{fig:preds_comparison} compares the confidence interval predictions from the Quantile-LSTM ensemble and the QDeepGR4J-LSTM ensemble for the Pascoe River at Fall Creek station (102101A) located in Queensland. We can observe that the median value predictions (red) from the hybrid model are closer to the observed values in all three time-steps in the prediction horizon. We also observe that the hybrid ensemble is better able to capture the streamflow peaks due to a wider confidence bound. However, some of the peaks are overestimated by the hybridised model.  Although hybridisation provides an overall improvement in the interval score, an overestimated peak could trigger false-positive alerts for flood warnings.

\subsection{Flood Risk Indicator}

We compute the flood risk indicator based on the flooding threshold ($\gamma$) identified using the GEV of the annual maximum streamflow, as shown in Equations  \ref{eq:gev} and \ref{eq:flood_risk_indicator}. 
%%%

We identify the value of $\gamma$ using the inverse of the CDF function computed for a $k-\text{year}$ flood recurrence interval. The dependence on the subjective value of flooding thresholds and the lack of any observations for flood classification make it challenging to evaluate the accuracy of the flood risk indicator. Therefore, we approach the evaluation by computing the flood risk indicators based on four different flood recurrence interval values, i.e., 3-year, 5-year, 7-year and 10-year.
We note that higher recurrence intervals, such as 25, 50 and 100 years, would be suitable for evaluation of extreme events. However, due to limitations with the training data length of approximately 25 years, the GEV threshold estimates for these recurrence intervals exceed the maximum observed flows. Consequently, no observed events were available for validation at these higher recurrence levels, and our analysis focuses on shorter recurrence intervals for flood risk validation.
We then use the  $p$ values for these flood recurrence intervals to compute the corresponding $\gamma$ values (Section \ref{sec:flood_risk_indicator}). Finally, we evaluate the accuracy of the flood risk indicator for the $\gamma$ values corresponding to the flood recurrence intervals by assigning binary flooding labels to both predicted streamflow quantiles and the observed streamflow. We assign the streamflow observations as a binary flooding label using an indicator transformation function $f = \mathbbm{1}(Q>\gamma)$. In the case of quantile-based predictions, we use the transformation function $\hat{f}=\mathbbm{1}(f(\hat{Q}_{0.05}) + f(\hat{Q}_{0.50}) + f(\hat{Q}_{0.95}) > 0)$  to assign a binary flooding label. We evaluate the flood risk indicator using the True Positive Rate (TPR), which is the ratio of the number of flood events correctly identified by the model (true positives) with respect to the total number of flooding events in the observations. Therefore, a higher value of TPR is desirable.

\begin{table*}
    \centering
    \begin{tabular}{lllrrrr}
        \toprule
                & & & \multicolumn{4}{c}{Flood Recurrence Interval}\\
                & & & \multicolumn{4}{c}{(years)} \\
        Station Id & Station Name & Model & 3 & 5 & 7 & 10 \\
        \midrule
        
        \multirow{2}{*}{116006B} & \multirow{2}{3cm}{Herbert River at Abergowrie} & LSTM & 0.926 & 0.000 & 0.000 & 0.000 \\
                & & DeepGR4J-LSTM & 0.963 & 1.000 & 0.750 & 1.000 \\
        \midrule
        
        \multirow{2}{*}{121001A} & \multirow{2}{3cm}{Don River at Ida Creek} & LSTM & 0.000 & 0.000 & 0.000 & 0.000 \\
                & & DeepGR4J-LSTM & 0.923 & 0.750 & 0.000 & 0.000 \\
        \midrule
        
        \multirow{2}{*}{122004A} & \multirow{2}{3cm}{Gregory river at Lower Gregory} & LSTM & 0.000 & 0.000 & 0.000 & 0.000 \\
                & & DeepGR4J-LSTM & 0.750 & 0.500 & 0.600 & 0.000 \\
        \midrule
        
        \multirow{2}{*}{126003A} & \multirow{2}{3cm}{Carmila Creek at Carmila} & LSTM & 0.000 & 0.000 & 0.000 &  \\
                & & DeepGR4J-LSTM & 0.800 & 0.625 & 0.333 &  \\
        \midrule
        
        \multirow{2}{*}{136202D} & \multirow{2}{3cm}{Barambah Creek at Litzows} & LSTM & 0.962 & 0.000 & 0.000 & 0.000 \\
                & & DeepGR4J-LSTM & 1.000 & 0.875 & 0.000 & 0.000 \\
        \midrule
        
        \multirow{2}{*}{137201A} & \multirow{2}{3cm}{Isis River at Bruce Highway} & LSTM & 0.000 & 0.000 & 0.000 & 0.000 \\
                & & DeepGR4J-LSTM & 1.000 & 1.000 & 1.000 & 0.600 \\
        \bottomrule
    \end{tabular}
    \caption{Flood risk indicator performance of LSTM and QDeepGR4J-LSTM ensembles computed on six stations located at the eastern coast of Queensland}
    \label{tab:flood_risk_res}
\end{table*}

\begin{figure}[htb]
    \centering
    \begin{subfigure}[b]{0.49\textwidth}
        \includegraphics[width=\linewidth]{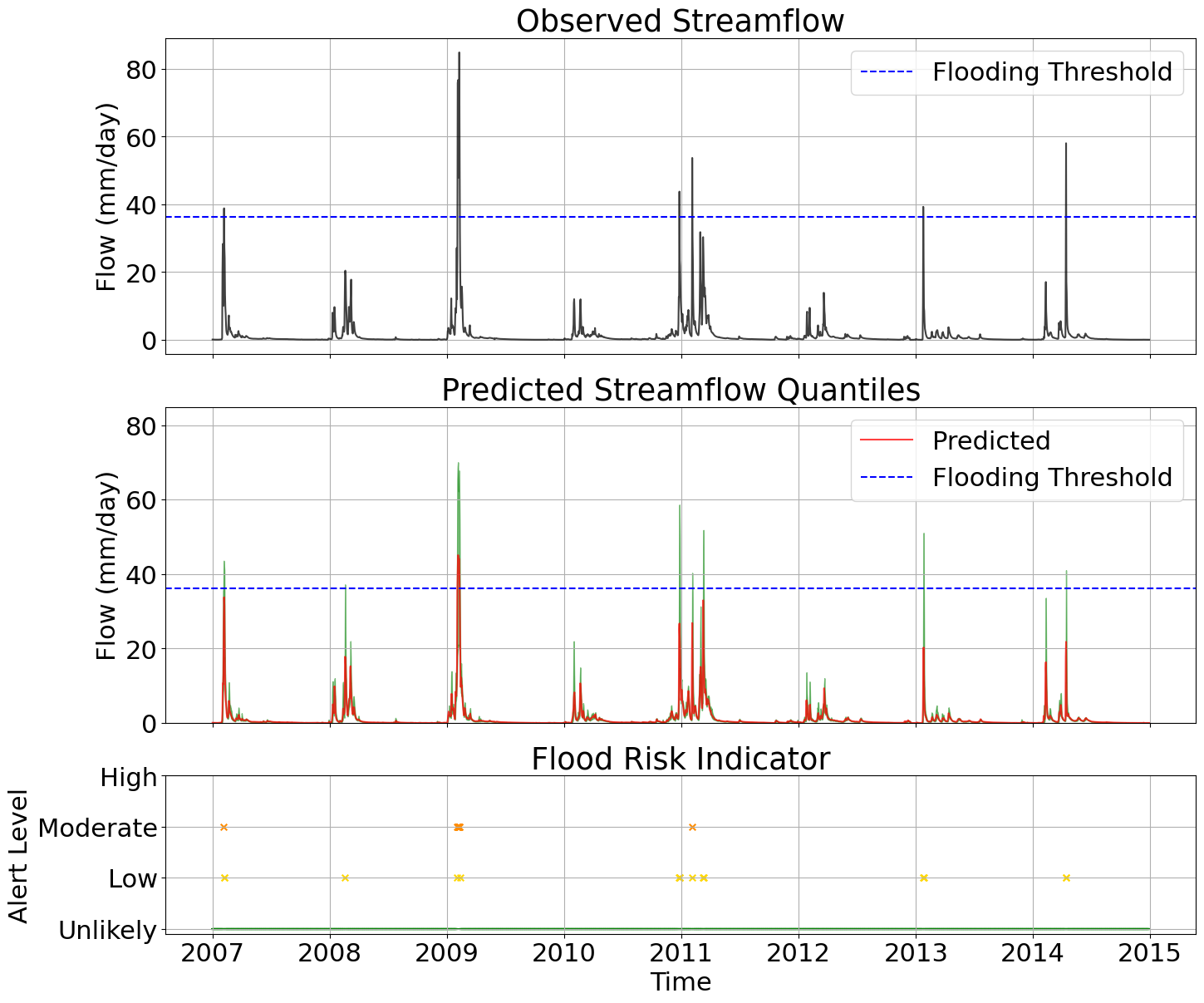}
        \caption{5-year flood recurrence interval}
    \end{subfigure}
    \begin{subfigure}[b]{0.49\textwidth}
        \includegraphics[width=\linewidth]{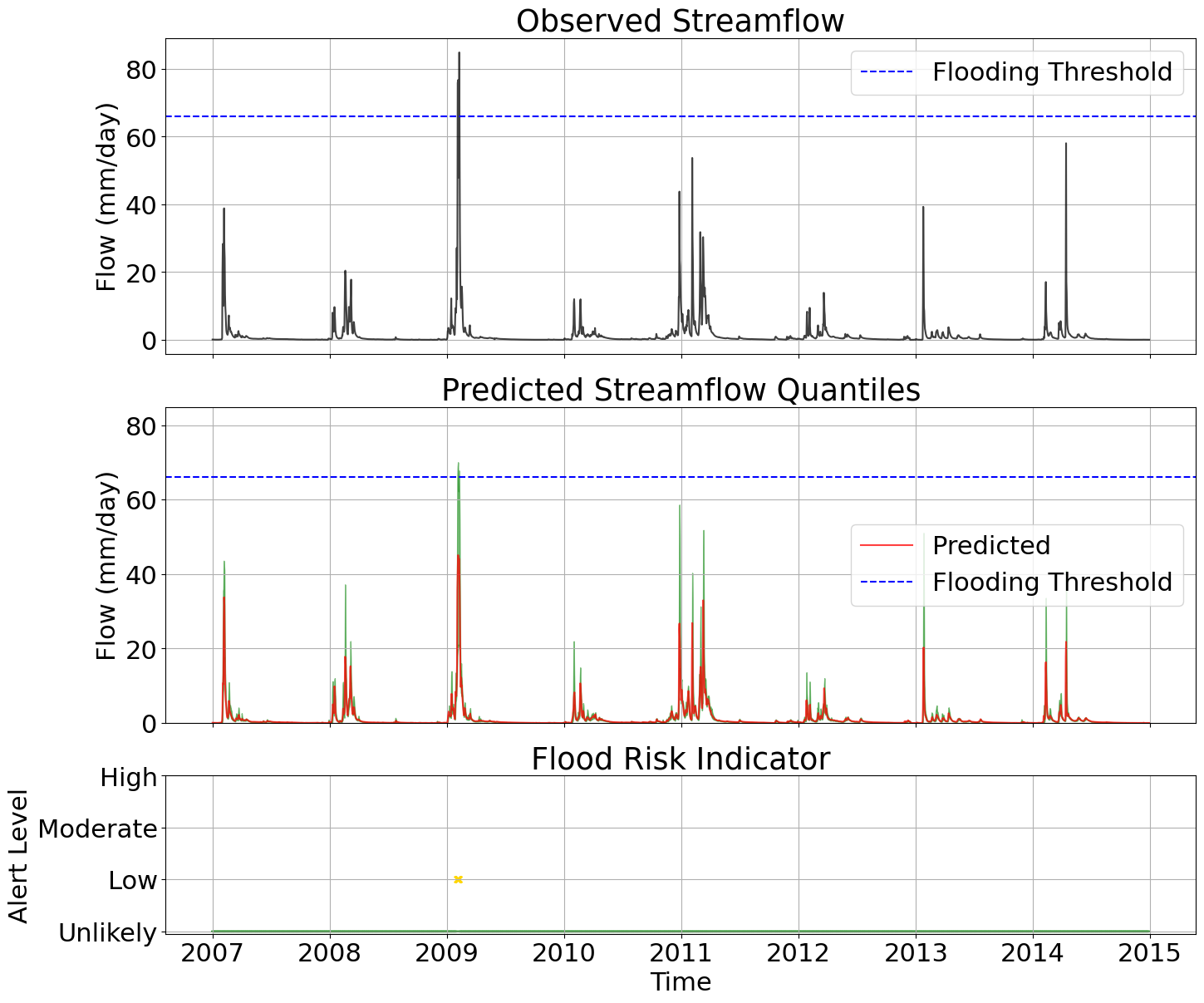}
        \caption{10-year flood recurrence interval}
    \end{subfigure}
    \caption{Flood risk indicator based on streamflow quantile predictions from LSTM-based QDeepGR4J ensemble on Herbert River at Abergowrie station (116006B)}
    \label{fig:flood_risk_res}
\end{figure}

Table \ref{tab:flood_risk_res} presents the TPR values of flood indicators computed using the LSTM ensemble and the DeepGR4J ensemble for six stations located at the south-eastern coast of Queensland. The results show that overall, the hybrid model shows a better performance in identifying the flooding events. The results show that the LSTM-based ensemble is unable to identify extreme events, particularly at higher values of flood recurrence intervals. The DeepGR4J-LSTM ensemble can capture almost all of the flooding events for a 3-year flood recurrence interval and close to half of the flooding events for a 5-year flood interval. However, for 7-year and 10-year floods, the TPR values are much lower, highlighting the limitations of this approach. Figure \ref{fig:flood_risk_res} shows a comparison of the flood risk indicators computed for 5-year flood recurrence interval and 10-year flood recurrence interval thresholds for the Herbert River at Abergowrie station (116006B) using the DeepGR4J-LSTM ensemble. In the 5-year flood recurrence interval, we observe that the model can capture most of the potential flooding events; however, we also notice that the model overestimates the streamflow (eg, early 2008) on occasion, leading to false positive alerts. In the case of a 10-year flood recurrence interval, the model can capture the high flow peak observed in the data.

%%%%%%%%%%%%%%%%%%%%%%%%%%%%%%%%%%%%%%%%%%%%%%%%%%%%%%%%%%%
% Discussion
%%%%%%%%%%%%%%%%%%%%%%%%%%%%%%%%%%%%%%%%%%%%%%%%%%%%%%%%%%%
 
\section{Discussion}

Our experimentation and results demonstrate the potential of a quantile regression-based hybrid rainfall model ensemble for predicting streamflow over multiple time steps with uncertainty quantification. We leveraged DeepGR4J, a deep learning based hybridisation of the GR4J rainfall runoff model, and proposed a Quantile regression-based ensemble of DeepGR4J models. We evaluated this approach in three stages: 1) identify the most suitable neural network architecture; 2) compare the performance of the hybrid ensemble against a pure machine learning based approach; and 3) use predicted uncertainty bounds to compute and evaluate the flood risk indicator. 

We evaluated the efficacy of various neural network architectures in predicting the 90\% streamflow uncertainty bounds on all stations in the South Australia region. Due to computational limitations, we restricted the evaluation to one state (region) using  South Australia, which has only nine stations. The results from the experiments show that the LSTM-based hybrid rainfall runoff model outperforms the vanilla RNN, CNN and MLP-based architectures. This is in contrast to our previous results on single-step ahead streamflow prediction (mean-value) using a hybrid rainfall runoff model, where the CNN-based model outperformed the LSTM-based counterpart \citep{kapoor2023deepgr4j}.  We attribute this difference to the nature of the task: quantile regression which emphasises tail behaviour and multi-step temporal dependencies, which are better captured by the LSTM model's ability to retain long-term memory of feature dynamics from the GR4J's components. By contrast, CNNs are more effective for short-term, localised pattern extraction and thus proved stronger in single-step mean prediction tasks. LSTM models are therefore well-suited to modelling temporal data with persistent dependencies, especially in the context of multi-step uncertainty prediction. In addition, the architectures used here differ from our earlier work, as we implement an encoder–decoder LSTM. Our choice is motivated by prior studies showing improved multi-step prediction accuracy \citep{chandra_evaluation_2021, wu2024review}, which has also led to better accuracy in our experiments.

Our results (Table \ref{tab:top5_res}) demonstrate the efficacy of the QDeepGR4J ensemble across all states in the Australian continent. Due to computational limitations similar to previous experiments, we restricted our evaluation to only five stations within each state. However, these stations were selected based on their observed runoff ratios. We selected stations with high runoff ratios, since a high value could indicate a higher chance of flash flooding within the catchment during high precipitation. The results confirmed that the LSTM-based QDeepGR4J ensemble performs the best across different states in terms of NSE, as well as interval score performance. The lower interval score values show that the hybridised LSTM ensemble is better able to capture the data uncertainty with tighter uncertainty bounds and the closest 90\% confidence interval. However, on plotting and comparing the uncertainty bounds generated by the two models, we observe some limitations of the proposed models. While the uncertainty bounds produced by the QDeepGR4J-LSTM ensemble are better able to capture the peaks, some peaks are highly overestimated, as shown in Figure \ref{fig:preds_comparison}. These limitations with overestimation of upper quantiles could be addressed in future work by adopting a multi-quantile training with explicit non-crossing constraints \citep{bondell_noncrossing_2010}, applying post-processing calibration via quantile matching  \citep{li_bias_2010}, and training with proper scoring rules such as the CRPS to balance sharpness and reliability \citep{hersbach_decomposition_2000}. Furthermore, we observed that the CNN-based QDeepGR4J ensemble yields a lower interval score for QLD and WA states. Albeit small, this difference indicates potential for regional variation in optimal architecture. We note that, similar to DeepGR4J, the QDeepGR4J ensemble also requires careful selection of model and optimiser hyperparameters. In our case, models trained with 7 time-steps (days) of input window using Adam optimiser with $\beta_1=0.89$ and $\beta_1=0.97$ gave the best performance. We note that z-score normalisation was used to normalise the input features as well as the targets for the neural network models.

We utilised the  \textit{flood risk indicator} as a qualitative measure of flood likelihood within the forecast horizon. Our results demonstrate that the QDeepGR4J ensemble outperforms the Quantile-based LSTM ensembles on the six selected stations located close to the eastern coast of Australia. We observe that while the QDeepGR4J ensemble yields notably high TPR values for 3-year and 5-year flood recurrence intervals, the performance drops for 7 and 10-year recurrence intervals. We also observe that the hybrid model can capture extreme events effectively, but some overestimations lead to false alarms. The results imply that our framework is useful as a more reliable early warning system, but requires calibration of thresholds to minimise false positives. 

We note that the subjective nature of the flooding threshold and the lack of availability of a flooding indicator in the observation data make it challenging to compute and evaluate a qualitative flood risk. Furthermore, due to their nature, extreme events are very few compared to non-extreme flow events, making standard classification metrics such as accuracy unreliable metrics for the evaluation of extreme event classification. Therefore, we rely upon the TPR, which measures how many extreme events were correctly identified by the model.

Beyond its potential for early flood warning, the proposed QDeepGR4J framework has broader implications for water resource management. The probabilistic estimates provided by the prediction intervals can help inform the infrastructure design for flood assessment under evolving climate considerations, in accordance with guidance in Australian Rainfall and Runoff (ARR 2019) \citep{ARR2019}. Furthermore, multi-day forecasts with quantified uncertainty can support pre-release decisions for dam operations, demonstrated by \citet{delaney_forecast_2020} through ensemble streamflow predictions at Lake Mendocino. Similarly, water allocation planning can benefit from seasonal forecasts with uncertainty, as this would improve the timing and consistency of allocation announcements, particularly increasing efficiency in agricultural practices \citep{kaune_benefit_2020}.

Despite the advantages of our framework, we also identify some key limitations in this study. Firstly, we observe that the uncertainty bounds can widen excessively over multi-step horizons. Therefore, with a high number of time-steps in the prediction horizon, we found a higher chance of false positives. Furthermore, the accuracy of predictions from the hybridised model is partially dependent on the GR4J calibration. Therefore, the errors in GR4J prediction have a significant influence on deep model inputs. Although the quantile regression approach can quantify the aleatoric uncertainty arising from the data, it cannot capture the epistemic uncertainty relating to the model architecture/parameters. 
A natural point of comparison here is Bayesian approaches to uncertainty quantification, such as MCMC/DREAM or GLUE \citep{vrugt_equifinality_2009,vrugt2009accelerating}, which provide theoretically rigorous posterior estimates of parameter and prediction uncertainty \citep{duc_verification_2018}. Despite their strength, Bayesian inference techniques are computationally intensive and may not be feasible for deep learning or large-scale applications. Our hybrid quantile-based ensemble model offers an efficient alternative for operational contexts, though future work could explore hybrid approaches that leverage Bayesian inference for parameter uncertainty alongside quantile regression for data-driven variability.
We note that a key limitation of our study is the relatively short training period length (25 years), which restricts our ability to evaluate very rare events such as 50 or 100-year floods. While these higher recurrence intervals are highly relevant for infrastructure design and long-term planning, reliable evaluation would require longer observational datasets or stochastic extensions. Our current analysis at 5 and 10-year recurrence levels provides operationally relevant benchmarks for near-term flood forecasting, whereas design-level applications will require future work with extended data sources.
Furthermore, advanced architectures such as the attention mechanisms \citep{vaswani2017attention} could be adopted for better temporal learning. In addition, deploying the QDeepGR4J ensemble in real-time forecasting environments and extending it to ungauged basins could significantly advance its utility for flood risk management under increasing climate variability.

%%%%%%%%%%%%%%%%%%%%%%%%%%%%%%%%%%%%%%%%%%%%%%%%%%%%%%%%%%%
% Conclusion and Future Work
%%%%%%%%%%%%%%%%%%%%%%%%%%%%%%%%%%%%%%%%%%%%%%%%%%%%%%%%%%%
\section{Conclusion}

We presented QDeepGR4J, a quantile-based ensemble of the DeepGR4J hybrid rainfall-runoff model that incorporates quantile regression and ensemble learning for uncertainty quantification and extreme flow prediction. Our proposed approach integrates the GR4J's production storage with a quantile regression-based deep neural network ensemble that targets specific quantiles for streamflow. This approach leverages the strengths of both conceptual hydrological models and data-driven architectures to improve the simulation of streamflow quantiles, particularly during high-flow and flood conditions. Furthermore, we enhance the framework for multi-step ahead forecasting and use the GEV distribution to derive the flood thresholds for a qualitative measure of flood risk based on the predicted quantiles.

The experimental results across various catchments in the CAMELS-AUS dataset demonstrate that LSTM-based QDeepGR4J ensembles consistently outperform baseline CNN and LSTM ensembles in both predictive accuracy (RMSE \& NSE) and uncertainty interval quality (interval score). Notably, the QDeepGR4J ensembles demonstrate an improved TPR performance for flood event detection, especially for 3-year and 5-year flood recurrence intervals. This makes QDeepGR4J a suitable candidate for early warning systems for flood events. Finally, the successful generalisation across multiple Australian states underscores the model’s adaptability to hydrogeological variations.

%%%%%%%%%%%%%%%%%%%%%%%%%%%%%%%%%%%%%%%%%%%%%%%%%%%%%%%%%%%
% Acknowledgements
%%%%%%%%%%%%%%%%%%%%%%%%%%%%%%%%%%%%%%%%%%%%%%%%%%%%%%%%%%%
\section*{Acknowledgements}

The authors would like to thank the Australian Government for supporting this research through the Australian Research Council’s Industrial Transformation Training Centre in Data Analytics for Resources and Environments (DARE) (project IC190100031). The authors also acknowledge the Katana High Performance Computing (HPC) cluster supported by the University of New South Wales for providing the computation resources to run the experiments (DOI: 10.26190/669X-A286).

%%%%%%%%%%%%%%%%%%%%%%%%%%%%%%%%%%%%%%%%%%%%%%%%%%%%%%%%%%%
% CODE
%%%%%%%%%%%%%%%%%%%%%%%%%%%%%%%%%%%%%%%%%%%%%%%%%%%%%%%%%%%
\section*{Software and Data Availability}

The data and open-source code can be accessed at the associated GitHub repository\footnote {\url{https://github.com/DARE-ML/DeepGR4J-Extremes}}.

%%%%%%%%%%%%%%%%%%%%%%%%%%%%%%%%%%%%%%%%%%%%%%%%%%%%%%%%%%%
% REFERENCES
%%%%%%%%%%%%%%%%%%%%%%%%%%%%%%%%%%%%%%%%%%%%%%%%%%%%%%%%%%%
\bibliographystyle{elsarticle-harv} 
\bibliography{references}

\end{document}